%% file: main.tex
\documentclass[10pt,twocolumn,letterpaper]{article}

\usepackage{iccv}              


\definecolor{iccvblue}{rgb}{0.21,0.49,0.74}
\usepackage[pagebackref,breaklinks,colorlinks,allcolors=iccvblue]{hyperref}

\def\paperID{14223} 
\def\confName{ICCV}
\def\confYear{2025}

\title{VGDFR: Diffusion-based Video Generation with Dynamic Latent Frame Rate}

\author{Zhihang Yuan$^{*~1~2}$\and Rui Xie$^{*~1~2}$\and Yuzhang Shang$^{3}$\and Hanling Zhang$^{1~2}$\and Siyuan Wang$^{1}$\and Shengen Yan$^{2}$\and Guohao Dai$^{4~2}$\and Yu Wang$^{1}$ \and \\ $^1$Tsinghua University ~~~~$^2$Infinigence AI~~~~
 $^3$Illinois Tech ~~~~$^4$Shanghai Jiao Tong University
}
\usepackage{graphicx}
\begin{document}
\maketitle

\vspace{-10pt}


\renewcommand{\thefootnote}{\fnsymbol{footnote}}
\footnotetext[1]{Equal contribution. }
\renewcommand{\thefootnote}{\arabic{footnote}}

\input{sec/0_abstract}    
\input{sec/1_intro}
\input{sec/2_relatedwork}

\input{sec/3_method}

\input{sec/4_experiment}
\input{sec/5_conclusion}
{
    \small
    \bibliographystyle{ieeenat_fullname}
    \bibliography{main}
}

\end{document}

%% file: sec/0_abstract.tex
\begin{abstract}

Diffusion Transformer(DiT)-based generation models have achieved remarkable success in video generation. However, their inherent computational demands pose significant efficiency challenges.
In this paper, we exploit the inherent temporal non-uniformity of real-world videos, and observe that videos exhibit dynamic information density, with high-motion segments demanding greater detail preservation than static scenes.
Inspired by this temporal non-uniformity, we propose \textbf{VGDFR}, a \textbf{training-free} approach for Diffusion-based Video Generation with Dynamic Latent Frame Rate. VGDFR adaptively adjusts the number of elements in latent space based on the motion frequency of the latent space content, using fewer tokens for low-frequency segments while preserving detail in high-frequency segments. 
Specifically, our key contributions are: (1) A dynamic frame rate scheduler for DiT video generation that adaptively assign frame rates for video segments.
(2) A novel latent-space frame merging method to align latent representations with their denoised counterparts before merging those redundant in low-resolution space. (3) A preference analysis of Rotary Positional Embeddings (RoPE) across DiT layers, informing a tailered RoPE strategy optimized for semantic and local information capture.
Experiments show that VGDFR can achieve a speedup up to 3$\times$ for video generation with minimal quality degradation. \href{https://github.com/thu-nics/VGDFR}{Project Link.}

\end{abstract}

%% file: sec/1_intro.tex
\section{Introduction}
\label{sec:intro}
\begin{figure}[htbp]
    \centering
    \includegraphics[width=\linewidth]{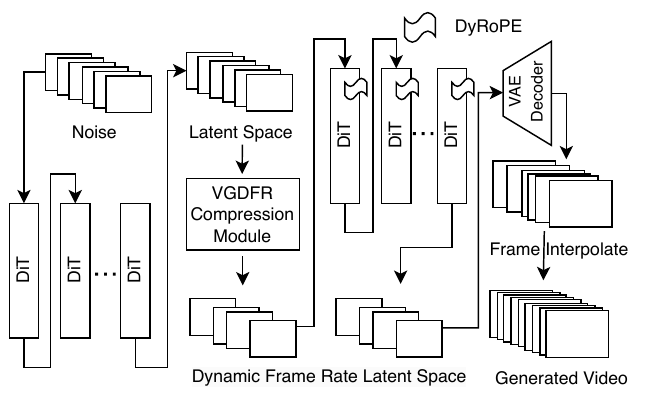}
    \caption{\textbf{VGDFR:} A training-free approach that accelerates video generation by dynamically determines the frame rate based on content already generated in denoising steps. By introducing VGDFR Compression Module, we apply various compression ratio of different time segments in order to reduce tokens for decreasing the computation cost of following DiT inference.}
    \label{fig:overview}
\end{figure}

Diffusion models (DMs) \cite{ho2020denoising,sohl2015deep} have achieved remarkable success in image synthesis \cite{rombach2022high,podell2023sdxl,saharia2022photorealistic} and video generation \cite{brooks2024video, kong2024hunyuanvideo,zheng2024open,girdhar2311emu, sepehri2024mediconfusion}, demonstrating exceptional performance in producing high-resolution and temporally coherent videos. However, their iterative denoising mechanism and the quadratic computational complexity of global attention \cite{vaswani2017attention} make high-resolution and long-duration video generation computationally expensive and inefficient in terms of inference speed. This bottleneck significantly limits practical applications; for instance, generating just a few seconds of video can take half an hour with a high-end GPU, making it challenging to meet the demands for longer and higher-quality video production.

Latent Diffusion Models (LDMs) \cite{rombach2022high} mitigate this issue by generating in a latent space. Each point in this latent space represents a fixed-size temporal and spatial area within the video. This significantly reduces computational requirements while preserving high generative performance. This approach has been adopted by Sora \cite{liu2024sora}, MovieGen \cite{polyak2025moviegencastmedia}, and HunyuanVideo \cite{kong2024hunyuanvideo}, all of which generate photorealistic videos.

Real-world videos exhibit significant temporal non-uniformity, with alternating high-motion and low-motion segments across time. High-motion segments require detailed representation to capture rapid changes, while low-motion segments contain redundant information that can tolerate higher compression. For instance, in the same video, the beginning part showing a static scene like ``a quiet library'' exhibits far greater compression potential than the latter part which is a dynamic segment depicting ``a person quickly moves into the library''.

Motivated by this, we propose \textbf{VGDFR}, a \textbf{training-free} approach for \textbf{Dynamic Latent Frame Rate Generation} in Diffusion Transformers (DiT). VGDFR compute the most suitable frame rate for each time segment online based on the motion frequency of the content, thereby enabling efficient video generation. VGDFR automatically calculates the optimal frame rate for each video segment, making it more adaptable to real-world generation scenarios. 
As shown in Figure~\ref{fig:overview}, we introduce a VGDFR Compression Module into the denoising process, which adaptively merges frames in the latent space. This allows subsequent denoising steps to be executed in a smaller latent space, significantly reducing computational costs.

In the VGDFR compression module, we use a dynamic frame rate scheduling strategy to divide the video into multiple time segments and assigns adaptive frame rates based on motion frequency and scene dynamics.
To determine the frame rates of different segments, we propose a novel latent space frame merging method. This method maps the latent space to its denoised counterparts and then decodes it into a low-resolution video space for merging redundant frames. In the low-resolution video space, we can conveniently calculate the similarity between different frames to identify low-motion segments, which can be compressed and utilize a lower latent frame rate.
For those low-motion segments, with the use of a lower latent frame rate, a single token can represent a longer duration of time. As a result, we can generate the video using a smaller number of tokens.
After compression, the number of tokens is reduced and their temporal relationships change. Therefore, the rotary positional encoding (RoPE) needs to be adjusted in the subsequent denoising steps. We dynamically adjust the parameters of rotary positional encoding (DyRoPE), taking into account the dynamic frame rate and the characteristics of different Transformer layers.

We conduct experiments on the HunyuanVideo~\cite{kong2024hunyuanvideo}. VGDFR can be easily applied to the HunyuanVideo DiT model without any fine-tuning. The results demonstrate that VGDFR effectively transforms the HunyuanVideo DiT into a network capable of generating videos in the dynamic latent space, significantly reducing computational costs. The transformed network achieves up to a $3\times$ speedup in video generation with minimal degradation in quality.




%% file: sec/2_relatedwork.tex
\section{Related Work}
\label{sec:relatedwork}

\textbf{Video Generation Model}. Early research in video generation primarily utilized Generative Adversarial Networks (GANs) \cite{NIPS2014_5ca3e9b1,karras2019style,tulyakov2018mocogan} and Variational Autoencoders (VAEs) \cite{kingma2013auto}. Despite their popularity, these models had notable limitations, particularly concerning generation quality and training stability. Recently, the significant progress of diffusion models in image generation tasks such as Stable-Diffusion~\cite{rombach2022high}, PixArt \cite{chen2023pixart} and Flux \cite{flux2024} has prompted researchers to explore their potential in video generation. Recently, a significant architectural shift has occurred within video diffusion models, transitioning from the previously dominant UNet-based frameworks to Transformer-based designs. Inspired by Diffusion Transformers (DiTs) \cite{peebles2023scalable, chen2023pixart}, this shift towards Transformers has allowed models to more effectively capture subtle temporal changes and complex motion patterns between frames, greatly enhancing generalization and detail recovery capabilities. As a result, video generation techniques~\cite{zheng2024open, lin2024open, kong2024hunyuanvideo} have advanced substantially, achieving higher precision and complexity.

Pixel-space diffusion models, although capable of generating high-quality outputs, typically demand substantial computational resources, limiting their widespread practical use. To overcome these challenges, Latent Diffusion Models (LDM) \cite{rombach2022high} were used. By performing the diffusion process in latent space, LDM considerably improve computational efficiency, lower hardware requirements, and thus accelerate the accessibility and application of video generation technology~\cite{blattmann2023stable,chen2024deep}. However, they use the same compression rate without considering that different segments of a video may have varying dynamic characteristics.

\textbf{Token Merge} is a promising technique aimed at reducing the computational complexity and memory overhead of vision Transformer models by merging similar tokens~\cite{bolya2022token}. Initially introduced to optimize Transformer architectures, this approach employs a similarity-based merging strategy to significantly decrease computational load while maintaining high accuracy. Recent works have applied Token Merging to accelerate image generation processes such as Stable Diffusion, achieving notable speed improvements without compromising visual quality \cite{bolya2023token, zhangtraining}. Transformer-based models often encounter substantial computational demands. To overcome this challenge, researchers have developed specialized variants of token merging designed for video, incorporating spatiotemporal token fusion within self-attention mechanisms. Such approaches have demonstrated effectiveness in compressing video representations, leading to reduced memory usage and improved computational efficiency, as well as enhanced temporal coherence in generated videos \cite{li2024vidtome,yuan2025dlfr,wang2025omnitokenizer}. However, they do not always guarantee generating the same content as the original videos. In contrast, VGDFR can perform token merging in the latent space and still generate almost identical video content.

%% file: sec/3_method.tex
\section{Method}
\label{sec:method}

In this section, we propose VGDFR, a training-free video generation method that dynamically determines the frame rate compression ratio based on the content in different time segments, enabling dynamic latent frame rate video generation.

We first propose our motivation by analyzing temporal frequency of latent space and feasibility of frame merging.
Based on this analysis, we propose the VGDFR compression module to compress the frames in the latent space to enable the dynamic frame rate in latent space. Additionally, we analyze the attention preferences of different Transformer layers, generating DyRoPE for them.

\subsection{Motivation}
\label{sec:motivation}


Traditional frame rate optimization in video processing and compression has primarily concentrated on raw video signals \cite{mackin2015study,song2001rate}. However, recent research demonstrates that operating in the latent space can yield results comparable to those achieved in the pixel space while significantly reducing computational overhead \cite{rombach2022high}.
Moreover, studies in video content analysis suggest that the information density of videos varies significantly over time~\cite{yuan2025dlfr}. For video segments with uneven information density, applying a uniform compression rate in the latent space can lead to considerable redundancy. For example, a high-motion segment such as \textit{a person rapidly moving into the library} contains far more dynamic information than a low-motion segment like \textit{a quiet library}. The latter should ideally be compressed at a higher rate. Building on this observation, dynamic frame rates can be employed in the latent space to enable adaptive compression ratios.


We aim to generate videos with dynamic frame rates that adapt to the video content over time. This approach involves compressing the latent space during the denoising process, which raises a critical question: can a latent segment be represented with different frame rates while preserving its semantic information? In other words, can a single token represent spacetime segments of varying temporal lengths without altering the meaning of the content?


To investigate this, we use a HunyuanVideo-VAE to test whether a latent segment can maintain its semantic consistency under different frame rates. Specifically, we process the same video content at different frame rates, compress it into the latent space using a VAE encoder, and then reconstruct it back into the original video space using the VAE decoder.
By comparing videos with different frame rates, we observe that motion in frame-compressed videos appears faster, but the start and end points, content, and motion amplitude remain consistent with the original video. This demonstrates that it is feasible to represent the original video content with fewer frames without losing essential information. Finally, by interpolating frames to restore the video to its original length, the resulting video is nearly indistinguishable from the original, further validating the approach.


\subsection{VGDFR Compression Module}
\label{sec:3.2}
\begin{figure*}[htbp]
    \centering
    \includegraphics[width=\linewidth]{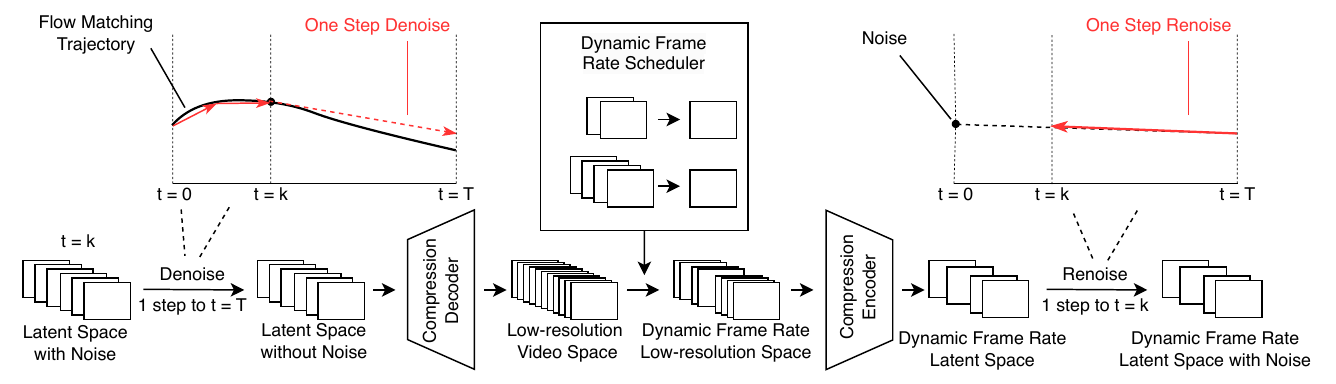}
    \caption{VGDFR Compression Module dynamically adjusts the frame rate for different time segments in a video to reduce the number of tokens needed for DiT inference. 
    }
    \label{fig:compression_module}
\end{figure*}

The goal is using fewer frames to represent the low-frequency motion segments, where the potential for token merging is vast. 
Some works employ QKV-based token merging techniques~\cite{sun2024asymrnr,zhang2024cross,zhao2024real,kahatapitiya2024adaptive}, which reduce the number of tokens involved in attention calculations. However, these methods need specific network architecture design or fine-tuning. 
Given the difficulty these methods face in adapting to the complexity of video generation, we propose a training-free paradigm. During DiT denoising steps, we directly reduce the number of latent space tokens through frame merging, rather than only reducing the number of QKV tokens involved in attention calculation. 

To determine whether a frame belongs to a low-motion or high-motion segment, we should focus on the inter-frame similarity.
As shown in Figure~\ref{fig:compression_module}, we introduce the VGDFR compression module to calculate the inter-frame similarity and then compress the similar frames.
There are five steps to achieve this.

\textbf{Step 1: One-step denoise}.

In video generation, the process begins by sampling Gaussian noise in the latent space. This initial noisy latent, denoted as $X^0$, is input into the DiT (Diffusion Transformer) network, which predicts the optimal flow direction for denoising at each timestep. In flow matching, the model directly learns a continuous vector field $f_\theta(X^t, t)$ that guides the latent from noise to data. The denoising process can be described by the following ordinary differential equation (ODE):
\begin{equation}
\frac{dX^t}{dt} = f_\theta(X^t, t)
\end{equation}
where $X^t$ is the latent representation at time $t$. The model is trained to match the flow between the noise and the data distribution across $T$ timesteps. We denote the latent representation $X^t$ at denoising timestep $t$, which has the shape $N_{latent} \times H_{latent} \times W_{latent}$. Here, $N_{latent}$ represents the temporal dimension, while $H_{latent}$ and $W_{latent}$ correspond to the height and width dimensions, respectively. Through this continuous transformation, the model gradually refines the noisy into a coherent video in latent space.

To address the lack of content information at $t=0$ where only pure Gaussian noise $X^0$ is available, we perform $k$ denoising steps to obtain a latent $X^k$ that contains preliminary video content. Since $k$ is set to a small value, $X^k$ retains a high noise level. For an efficient early preview of the video content, we adopt a one-step denoising strategy that directly maps $X^k$ to the final clean latent $X^T$. This can be formulated as:
\begin{equation}
X^T_{pre} = X^k + f_\theta(X^t, t) \frac{T-k}{T}
\end{equation}
Here, the model predicts the integrated flow from step $k$ to $T$ in a single iteration, effectively removing most of the noise and revealing the coarse structure of the generated video. This rapid approximation enables us to quickly visualize the video’s overall content during the early stages of generation.

\textbf{Step 2: Compression Decoder}.

After obtaining $X^T_{pre}$, we note that it resides in the latent space, where the data distribution is highly complex. Unlike the pixel space, it is difficult to directly assess frame-to-frame differences in the latent domain. Therefore, we aim to project the latent back to the pixel space. However, using the original VAE Decoder is computationally expensive, primarily due to the spatial upsampling operations, which significantly slow down the VGDFR compression module.

Since our goal is to distinguish high-motion and low-motion video segments, a high-resolution reconstruction is unnecessary. Therefore, we propose a compression decoder, modified from the original VAE Decoder by removing most of the spatial upsampling layers. This design greatly improves efficiency—achieving more than a 16$\times$ speedup—while retaining enough content information for motion analysis.
Formally, given the latent $X^T_{pre}$, the compression decoder $D_{comp}$ outputs a low-resolution video $Y$ as follows:
\begin{equation}
Y_{low} = D_{comp}(X^T_{pre})
\end{equation}
where $Y_{low}$ has the shape $N_{low} \times H_{low} \times W_{low}$. This efficient low-resolution preview enables fast motion analysis for subsequent video segmentation tasks.

\textbf{Step 3: Dynamic Frame Rate Schedule}.

Based on the low-resolution video $Y_{low}$, we can decide whether to merge frames within a video segment. Specifically, we introduce a threshold $\theta$ to control the merging process. Let $S_{i,j} = \text{SIM}(Y_{i}, Y_{j})$ denote the similarity between the $i$-th and $j$-th frames.
A video segment is eligible for merging if and only if the similarity between every pair of frames within the segment (frame index $I$ to $J$) exceeds the threshold $\theta$:
\begin{equation}
S_{i,j} > \theta, \quad \forall\, i, j \in [I,J]
\end{equation}
If this condition is satisfied, all frames in the segment are merged by taking their mean.
This dynamic frame rate schedule enables adaptive temporal compression, preserving motion details in high-motion regions while efficiently merging redundant frames in low-motion segments.

\textbf{Step 4: Compression Encoder}.

Similar to the Compression Decoder, we modify the original VAE Encoder by removing most of the downsampling operations to create the Compression Encoder. This modification allows the encoder to efficiently map the compressed low-resolution video back to the latent space. Let $Y_{dy,low}$ represent the low-resolution video with dynamic frame rate obtained from Step 3. The Compression Encoder $E_{comp}$ performs the following mapping:
\begin{equation}
X_{dy,pre}^T = E_{comp}(Y_{dy,low})
\end{equation}
where $X_{dy,pre}^T$ is the compressed latent representation with dynamic frame rate. By bypassing unnecessary downsampling layers, this encoder ensures that the compressed low-resolution video can be efficiently mapped to a compact latent space, enabling faster computations and reduced memory usage.

\textbf{Step 5: Renoise}.

To resume denoising in the compressed latent space, we first reconstruct the noisy latent at timestep $k$ from the compressed clean latent $X_{dy,pre}^T$ and the original noise $X^0$ via linear interpolation. This operation aligns with the flow matching paradigm, where denoising follows a continuous trajectory from noise to data.
We select the corresponding frames from the original noise $X^0$ according to the compressed temporal indices, resulting in $X^0_{dy}$. 
Then, we compute the re-noised latent $X_{dy}^k$ as:
\begin{equation}
X_{dy}^k = \left(1 - \frac{k}{T}\right) X^0_{dy} + \frac{k}{T} X_{dy,pre}^T
\end{equation}

This interpolation ensures that $X_{dy}^k$ lies on the same trajectory between the original noise and the compressed clean latent, consistent with the continuous vector field learned by the model. Once $X_{dy}^k$ is obtained, we resume denoising in the compressed latent space from timestep $k$ to $0$, completing the generation process efficiently under the dynamic frame rate schedule.

\textbf{Denoise in Compressed Latent Space}.

After obtaining $X_{dy}^k$, we apply the DiT model to perform denoising from timestep $k$ to $T$ within the compressed latent space. This results in a clean latent representation $X_{dy}^T$ adapted to the dynamic frame rate schedule.

We then use the VAE Decoder to map $X_{dy}^T$ back to the pixel space and reconstruct the final video. However, due to the temporally compressed nature of the latent sequence, certain frames have been merged, and thus we need to restore the full frame rate. To achieve this, we use an interpolation network capable of generating the missing frames between two decoded frames while maintaining temporal coherence and motion consistency~\cite{huang2022rife,huang2024safa}.

Formally, given two consecutive decoded frames $Y_i$ and $Y_{i+1}$ with a time interval $\Delta t > 1$, the interpolation network $\mathcal{I}$ predicts intermediate frames $\{Y_{i+\delta}\}_{\delta=1}^{\Delta t - 1}$ as:
\begin{equation}
Y_{i+\delta} = \mathcal{I}(Y_i, Y_{i+1}, \delta, \Delta t), \quad \text{for } \delta = 1, \dots, \Delta t - 1
\end{equation}

Through these steps, we are able to reconstruct high-quality videos that preserve both spatial fidelity and temporal dynamics, while benefiting from the efficiency of latent space compression and adaptive frame sampling.

\begin{table*}[t]
\centering
\caption{\textbf{Performance of VGDFR text-to-video generation on the VBench benchmark.} The parameter $''k''$ indicates the starting timestep for dynamic frame rate compression, while $''\theta''$ represents the inter-frame similarity threshold.}
\label{tab:exp}
\begin{tabular}{@{}cccccccccc@{}}
\toprule
                       & \multicolumn{4}{c}{Video Quality Metrics}            & \multicolumn{3}{c}{Difference Metrics} & \multicolumn{2}{c}{Efficiency} \\ \cmidrule(l){2-10} 
Method                 & CLIPT       & CLIPSIM     & VQA         & FLICKER    & PSNR         & SSIM       & LPIPS      & Latency(s)        & Speedup       \\ \midrule
Original               & 0.9994   & 0.1871    & 98.63   & 0.9793   & -            & -          & -          & 714.90         & -             \\ \midrule
VGDFR (k=5, $\theta$=0.5)  & 0.9996      & 0.1861      & 97.85       & 0.9919     & 15.399       & 0.513      & 0.398      & 245.12         & 2.92          \\
VGDFR (k=5, $\theta$=0.6)  & 0.9995      & 0.1859      & 98.26       & 0.9912     & 15.63        & 0.525      & 0.383      & 282.02         & 2.53          \\
VGDFR (k=5, $\theta$=0.7)  & 0.9995      & 0.1859      & 98.16       & 0.9903     & 15.784       & 0.531      & 0.380       & 299.96         & 2.38          \\
VGDFR (k=5, $\theta$=0.8)  & 0.9995      & 0.1863      & 97.92       & 0.9886     & 16.463       & 0.566      & 0.348      & 400.77         & 1.78          \\
VGDFR (k=5, $\theta$=0.9)  & 0.9996      & 0.1857      & 97.58       & 0.9865     & 17.722       & 0.631      & 0.293      & 547.32         & 1.31          \\ \midrule
VGDFR (k=10, $\theta$=0.5) & 0.9996      & 0.1851      & 97.31       & 0.9912     & 17.348       & 0.574      & 0.326      & 306.56         & 2.33          \\
VGDFR (k=10, $\theta$=0.6) & 0.9995      & 0.1852      & 97.44       & 0.9907     & 17.47        & 0.581      & 0.319      & 325.81         & 2.19          \\
VGDFR (k=10, $\theta$=0.7) & 0.9995      & 0.1856      & 97.52       & 0.9891     & 17.73        & 0.597      & 0.308      & 359.06         & 1.99          \\
VGDFR (k=10, $\theta$=0.8) & 0.9995      & 0.1854      & 97.69       & 0.9871     & 18.531       & 0.638      & 0.277      & 445.3          & 1.61          \\
VGDFR (k=10, $\theta$=0.9) & 0.9995      & 0.1862      & 97.42       & 0.9859     & 19.639       & 0.691      & 0.237      & 562.69         & 1.27          \\ \midrule
VGDFR (k=15, $\theta$=0.5) & 0.9996      & 0.1843      & 96.57       & 0.9907     & 18.438       & 0.607      & 0.293      & 364.45         & 1.96          \\
VGDFR (k=15, $\theta$=0.6) & 0.9995      & 0.1846      & 96.54       & 0.9899     & 18.587       & 0.617      & 0.286      & 378.72         & 1.89          \\
VGDFR (k=15, $\theta$=0.7) & 0.9995      & 0.1845      & 96.78       & 0.9883     & 18.877       & 0.637      & 0.273      & 408.58         & 1.75          \\
VGDFR (k=15, $\theta$=0.8) & 0.9995      & 0.1850       & 96.95       & 0.9859     & 19.724       & 0.681      & 0.240       & 493.33         & 1.45          \\
VGDFR (k=15, $\theta$=0.9) & 0.9995      & 0.1854      & 97.09       & 0.9853     & 20.606       & 0.723      & 0.210       & 577.28         & 1.24          \\ \bottomrule
\end{tabular}
\end{table*}

\subsection{Dynamic Rotary Position Embedding}
\label{sec:rope}

Video generation utilizes RoPE (Rotary Position Embedding) to represent positional relationships between different tokens. Each token is associated with a set of sine and cosine values derived from RoPE to encode its position, with separate channels for the height ($H_{latent}$), width ($W_{latent}$), and time ($N_{latent}$) dimensions. In our approach, we dynamically select frame rates and perform frame merging in the low-resolution video space, leading to non-uniform temporal intervals between frames. Consequently, the original RoPE encoding must be adjusted to reflect the updated temporal structure.

To address this, we first propose \textbf{Global-RoPE}. Based on the indices of merged frames recorded by the dynamic frame rate scheduler, we directly remove the corresponding RoPE components in the temporal dimension. This ensures that the remaining positional encodings focus solely on effective frames, eliminating the influence of redundant ones. For example, given frames $[1, 2, 3, 4]$, if frames $1$ and $2$ are merged, we retain only the RoPE values for frames $[1, 3, 4]$.

However, applying Global-RoPE across all DiT layers significantly degrades video quality. This is because dynamic frame rate selection alters the temporal distribution of tokens, diverging from the distribution the model was originally pre-trained on. Overusing Global-RoPE disrupts the model’s learned positional priors, resulting in poor generation quality.

To alleviate this issue, we introduce \textbf{Local-RoPE}. Instead of discarding RoPE components, we retain positional encodings for a continuous sequence of frames with the same length as the remaining frames. This helps preserve the structural prior of the original data while still enabling frame compression. For instance, if the original frames are $[1, 2, 3, 4]$ and frames $1$ and $2$ are merged, we apply RoPE values from frames $[1, 2, 3]$, maintaining a continuous encoding span and preserving temporal consistency.

We refer to this overall strategy as \textbf{Dynamic Rotary Position Embedding (DyRoPE)}, which adopts an adaptive design: layers that are more sensitive to distribution shifts use Local-RoPE, while layers that are more vulnerable to redundant information adopt Global-RoPE. This hierarchical adjustment enables layer-wise RoPE adaptation according to functional roles in the DiT model, leading to improved semantic understanding and higher-fidelity video synthesis.

%% file: sec/4_experiment.tex
\section{Experiments}
\label{sec:experiments}

\begin{figure*}[htbp]
    \centering
    \includegraphics[width=\linewidth]{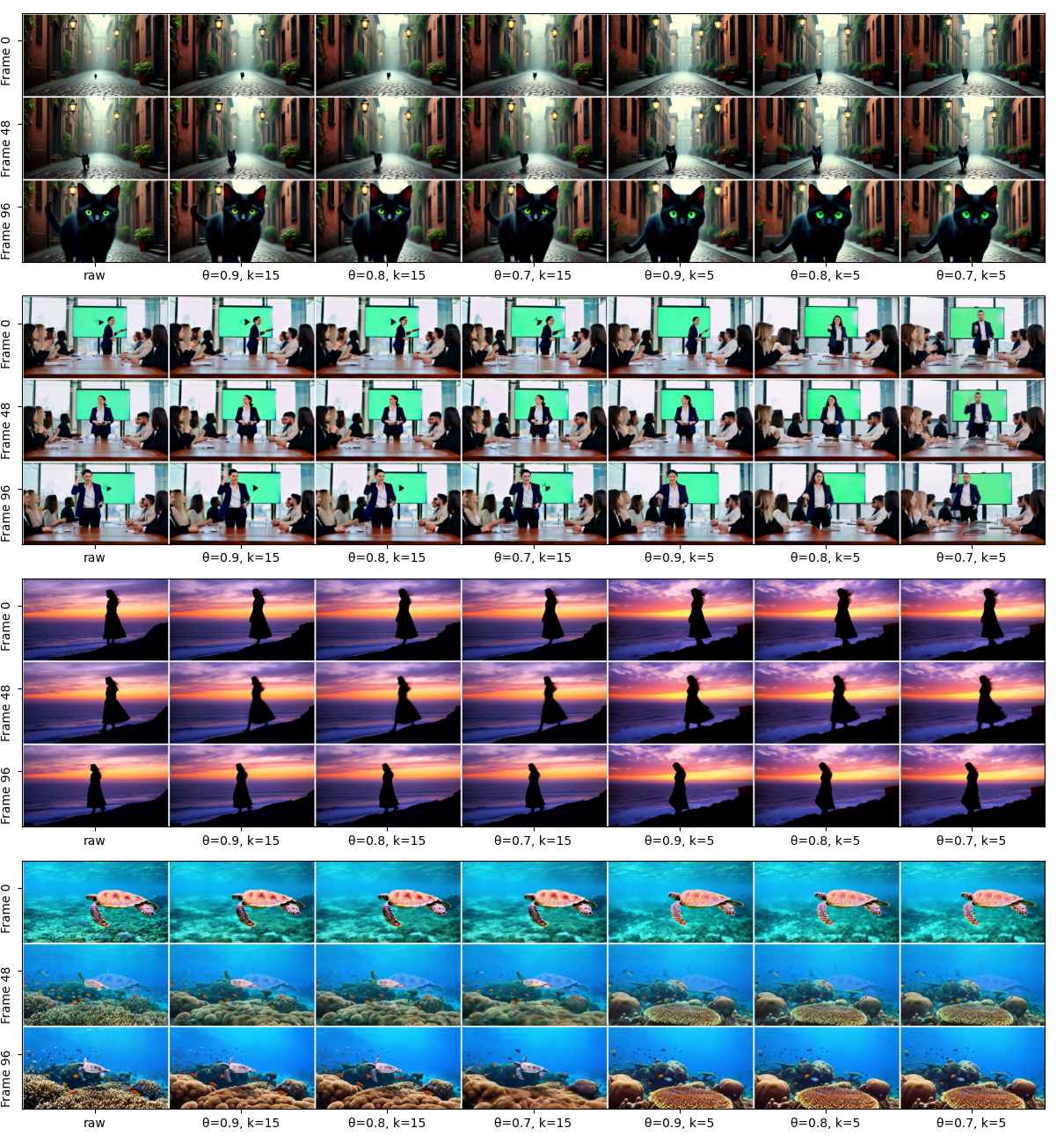}
    \caption{\textbf{Visualization of Text-to-Video Generation on VBench Prompt Set.} Three rows of the figure displays frames extracted from the start, middle, and latter of the generated videos. The leftmost column shows videos generated by the original Hunyuan Video model, while the subsequent columns present videos generated by VGDFR under different parameter configurations. Here, $\theta$ denotes the threshold, and $k$ represents the denoising timestep at which dynamic frame rate compression is initiated.}
    \label{fig:visualization}
\end{figure*}

\subsection{Experimental Settings}

Video Generation: We apply VGDFR to the Hunyuan  Video\cite{kong2024hunyuanvideo} model to generate 540P videos with 97 frames. For the RoPE configuration, we apply global RoPE to layers with indices [4, 19, 23, 31, 35, 36, 37, 40], while local RoPE is used for the remaining layers. We use RIFE 4.26 Large~\cite{huang2022rife,huang2024safa} to interpolate the video frames after VAE Decoder. We use SSIM to compute the similarity between different video frames. All experiments are performed on A100 80G GPUs to evaluate the generation quality and computational latency.

To validate the generation results, we utilize a subset of prompts collected from VBench \cite{huang2024vbench}. Similar to prior work \cite{xi2025sparse}, in order to avoid the destruction of the overall structure caused by compression during the early stages of denoising, we apply our dynamic frame rate merging method after the initial 10\%,20\%,30\% of the denoising steps.

\noindent \textbf{Evaluation Metrics:} We employ two types of metrics:

\noindent (1) Quality Metrics: These metrics assess the absolute quality of videos from various perspectives. We utilize CLIPSIM \cite{wu2021godiva} and CLIP-Temp \cite{esser2023structure} to evaluate text-video alignment and temporal consistency, respectively. We also use VQA~\cite{wu2022fastvqa} to evaluate the aesthetic score of the generated videos. Additionally, we utilize the temporal flickering score provided by VBench \cite{huang2024vbench} to measure temporal consistency at local and high-frequency details of generated videos. We
calculate the average MAE(mean absolute difference) value between each frame.
\noindent (2) Relative Difference Metrics: These metrics quantify the differences between videos generated using our method and those produced by the original Hunyuanvideo model. PSNR and Cosine Similarity are used to measure low-level pixel space differences. SSIM \cite{wang2004image} is employed to evaluate structural similarity, while LPIPS \cite{zhang2018unreasonable} is utilized to assess feature-space differences in the input data.

\subsection{Generation Results}

\textbf{Text-to-video generation: }
In Table~\ref{tab:exp}, we present the evaluation metrics. For different threshold values $\theta$, we employed three distinct starting timesteps for dynamic frame rate compression: $k=5,10,15$ (with a total denoising iteration timestep of 50). From the perspective of quality metrics, the videos generated by our method demonstrate quite promising results. Although the performance in similarity metrics is not entirely satisfactory, an analysis from the semantic and content levels reveals that our generated videos maintain a high degree of consistency with those produced by the original Hunyuan Video model. The primary reason for the lower similarity lies in the denoising process, which projects the video into a low-resolution space, followed by dynamic frame rate compression and then returns to the latent space for re-noising. This encoding-decoding process inevitably introduces noise discrepancies, making it hardly to generate completely same videos (for instance, the gender of speaker in the generated video changes, which is not specified by prompt, as shown in the middle video in Fig.\ref{fig:visualization}. However, these videos fully comply with the prompt in terms of content and maintain visual consistency with the videos generated by the original model. Our method achieves a speedup effect ranging from $1.2\sim3\times$. The latency of the original video is approximately $715$ seconds, while our method reduces this latency to as fast as $245$ seconds and at most $577$ seconds.

Additionally, we provide a qualitative demonstration in Fig.\ref{fig:visualization} showcasing a challenging scenario where there is a significant difference in the speed of the subject's movement between the earlier and later segments of the video (e.g., the cat walks toward the camara, starts slow and then accelerate).

\subsection{Ablation Study}
\begin{figure}[t]
    \centering
    \includegraphics[width=\linewidth]{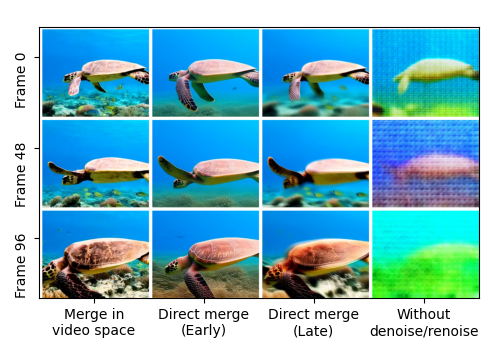}
    \caption{\textbf{Contribution of components.} Left column: Video generated by VGDFR. Middle two column: Video generated by VGDFR with Compression Module completely removed. Right column: Video generated by omitting denoising and renoising operations in the VGDFR Compression Module.}
    \label{fig:ablation}
\end{figure}


We analyze the contributions of different components of VGDFR, as illustrated in Fig.\ref{fig:ablation}. 

A comparison between the middle two columns and the left column reveals that omitting the compression module and operating directly in the latent space leads to a noticeable degradation in video quality. Specifically, initiating dynamic frame rate compression at earlier denoising timesteps leads to the loss of local details. This is because in earlier timesteps, where the content is predominantly noise, compression reduces the variance, thereby diminishing the intensity of the noise. Conversely, applying this compression at later timesteps results in artifacts such as blurry and regression. This phenomenon occurs because the later timesteps have already generated clearer content, leading to the blending of inter-frame information when compressing the frame rate.

Comparing the rightmost column with the leftmost one, we find that removing the denoising and renoising steps results in a complete breakdown of the generated video. This is due to our compression module utilizing pre-trained weights from a Variational Autoencoder (VAE), which has not been exposed to distributions with noisy latents.

%% file: sec/5_conclusion.tex
\section{Conclusion}
\label{sec:conclusion}
VGDFR, a training-free method based on Diffusion Transformer (DiT) model, enhances the efficiency of video generation by dynamically compressing the latent frame rate during denoising iteration. This approach fully leverages the inherent temporal non-uniformity of real-world videos, where the content exhibits dynamic motion variation frequency across different time segments. By preserving more details in high-motion-frequency segments and reducing redundancy in low-motion-frequency segments, VGDFR significantly improves video generation speed while maintaining generation quality.

Experimental results demonstrate that VGDFR achieves a $1.2\sim3\times$ speed up in generation without additional training or tuning, while maintains high levels of content and semantic consistency with original DiT's generation. This method introduces a novel, training-free, and efficient paradigm for video generation, substantially reducing computational overhead and enabling the production of high-quality videos.